\pdfoutput=1
\documentclass[11pt]{article}

\usepackage[T1]{fontenc}
\usepackage[utf8]{inputenc}
\usepackage{lmodern}
\usepackage[margin=1in]{geometry}
\usepackage{microtype}
\usepackage{amsmath,amssymb}
\usepackage{booktabs}
\usepackage{tabularx}
\usepackage{array}
\usepackage{xcolor}
\usepackage{graphicx}
\usepackage{setspace}
\usepackage{caption}
\usepackage{titlesec}
\usepackage{authblk}
\usepackage{placeins}
\usepackage{url}
\usepackage[super,square,comma]{natbib}
\usepackage[colorlinks=true,linkcolor=blue!55!black,citecolor=blue!55!black,urlcolor=blue!55!black]{hyperref}

\hypersetup{
  pdftitle={EPPC-CARE: Staged Hierarchical Annotation of Communication Behaviors in Secure Patient--Provider Messages},
  pdfauthor={Samah Fodeh, Sreeraj Ramachandran, Ganesh Puthiaraju, Sarah Schellhorn},
  pdfkeywords={Patient Portals, Natural Language Processing, Large Language Models, Information Extraction, Professional--Patient Relations}
}

\titleformat{\section}{\large\bfseries}{\thesection}{0.75em}{}
\titleformat{\subsection}{\normalsize\bfseries}{\thesubsection}{0.75em}{}
\titleformat{\subsubsection}{\normalsize\itshape\bfseries}{\thesubsubsection}{0.75em}{}

\makeatletter
\def\input@path{{drafts/}{figures/}}
\makeatother
\graphicspath{{figures/}}

\title{\Large\bfseries EPPC-CARE: Staged Hierarchical Annotation of Communication Behaviors in Secure Patient--Provider Messages}

\author[1,*]{Samah Fodeh, PhD}
\author[1]{Sreeraj Ramachandran, PhD}
\author[1]{Ganesh Puthiaraju, PhD}
\author[2]{Sarah Schellhorn, MD}

\affil[1]{Department of Emergency Medicine, Yale School of Medicine, New Haven, CT, USA}
\affil[2]{Medical Oncology, Yale School of Medicine, New Haven, CT, USA}
\affil[*]{Corresponding author: Samah Fodeh, PhD, Department of Emergency Medicine, Yale School of Medicine, 300 George Street, New Haven, CT 06511, USA. Email: \texttt{samah.fodeh@yale.edu}}

\date{}

\begin{document}

\maketitle
\vspace{-1.2em}

\begin{center}
\rule{\textwidth}{0.5pt}
\end{center}
\noindent{\large\bfseries Abstract}
\vspace{0.4em}

{\small
\noindent\textbf{Objectives:} Messages in the patient portal provide invaluable information about the dynamics of communication between patients and their healthcare providers, but extracting this information manually is labor-intensive. We developed and evaluated EPPC-CARE, an automated framework for identifying communication behaviors together with the supporting text for each classification.

\medskip
\noindent\textbf{Materials and Methods:} From 867 messages, we obtained 5,516 reference standard annotations. Each annotation is a triplet of communication-related text (span), a broader communication category (Code), and a specific subcategory (Sub-code). We evaluated EPPC-CARE's identified triplets across six open-weight large language models (LLMs) against supervised fine-tuning and other communication-based models such as EPPC-OASIS. We also compared EPPC-CARE with human expert annotations on 100 messages.

\medskip
\noindent\textbf{Results:} EPPC-CARE improved performance over EPPC-OASIS across all six models. EPPC-CARE's Triplet F1 improved by 2.90 percentage points on average, with a best score of 64.4\%. On the human expert evaluation subset, EPPC-CARE and the human expert showed comparable agreement with the reference standard, with differences favoring EPPC-CARE.

\medskip
\noindent\textbf{Discussion:} These findings suggest that identifying the triplet components (i.e., span, code, subcode) sequentially can improve performance across model families while preserving the supporting text behind each identification. EPPC-CARE links the communication code and subcode with the source text (span), supporting human verification of automated annotations in clinical informatics. EPPC-CARE may therefore make larger-scale studies of patient--provider communication more feasible and provide a foundation for human-in-the-loop annotation workflows.

\medskip
\noindent\textbf{Conclusion:} EPPC-CARE offers a more reliable and reviewable approach for automated characterization of communication behaviors in secure patient--provider messaging.

}

\begin{center}
\rule{\textwidth}{0.5pt}
\end{center}

\section{Background and Significance}\label{sec:introduction}

Secure messaging has become a routine part of care, allowing patients and clinical teams to exchange information and coordinate care outside scheduled encounters.\cite{Antonio2020PortalEvidence,Dendere2019PortalReview} These messages capture questions, explanations, requests, logistical needs, expressions of concern, and efforts to reach shared decisions; features of patient--provider communication that are not readily represented by diagnoses, procedures, or other structured clinical data.\cite{Cronin2015SecureMessaging,Huang2022SecureMessages} This makes secure messages a valuable source for studying how patients and clinical teams communicate. However, the growing volume of messaging makes manual review and coding increasingly difficult for studies conducted at scale.\cite{North2020SecureMessages} Automated methods that can identify communication behaviors while retaining the supporting text for each identification could enable larger studies without losing the ability to inspect the underlying language.

The Electronic Patient--Provider Communication (EPPC) framework provides a structured approach for characterizing these communication behaviors.\cite{Fodeh2026EPPCMinerBen} Each behavior is represented by a broad communication category (Code), a specific communication subcategory within that category (Sub-code), and the exact supporting text from the message (Span). For example, communication about an appointment may be assigned the Code \emph{Information Giving} and the Sub-code \emph{Appointment Related Information}, with the relevant words in the message retained as the Span.\cite{Fodeh2026EPPCMinerBen} A single message may contain several communication behaviors and therefore several triplets. This structure preserves a direct connection between each classification and the language on which it is based, but it also makes automated annotation challenging. A system must determine how many behaviors are present, identify the Span for each behavior, and assign the corresponding Code and Sub-code. These decisions can be particularly difficult when messages contain multiple or overlapping communication functions expressed in informal language.

Large language models (LLMs) offer a potential way to automate this process because they can interpret free text and produce a variable number of structured annotations.\cite{Agrawal2022ClinicalIE,Builtjes2025OpenIE} A single-pass approach, however, requires the model to identify the Span and assign both the Code and Sub-code simultaneously. Errors in one part of this process can affect the others: an incorrect Span may lead to the wrong classification, while misinterpreting the communication behavior may lead the model to focus on the wrong portion of the message. Making all of these decisions at once also makes it more difficult to determine where an incorrect triplet originated.

EPPC-CARE addresses this problem by dividing the annotation task into three sequential decisions. It first identifies candidate Spans, then refines them while assigning Codes, and finally selects valid Sub-codes. EPPC-CARE also retrieves semantically similar annotated messages to provide task-relevant examples during this process. This design differs from existing methods that refine candidate annotations containing all three components together.\cite{Fodeh2026EPPCOASIS,fodeh2026stardro,Fodeh2026EPPCMinerBen} By constructing the annotation incrementally, EPPC-CARE is intended to reduce the number of decisions made simultaneously while maintaining a traceable connection from the source text to the final classification.

\subsection{Objective}\label{sec:objective}

To support scalable and reviewable analysis of patient--provider communication, we developed and evaluated EPPC-CARE, a staged approach for automatically identifying communication behaviors together with their Spans. We designed EPPC-CARE to separate Span identification, Code assignment, and Sub-code selection rather than requiring all three decisions to be made simultaneously. We evaluated whether this approach improved triplet performance compared with supervised fine-tuning and other existing models across six open-weight LLMs spanning multiple model families and sizes. We further assessed whether performance gains were consistent across models and compared EPPC-CARE and a human expert with the reference standard to examine their comparative performance.

\subsection{Related work}\label{sec:related}

Secure patient--provider messaging has been studied as both a source of information about care communication and a growing component of clinical workload.\cite{Cronin2015SecureMessaging,North2020SecureMessages,Huang2022SecureMessages,Hoonakker2017SecureWorkflow,Arndt2017EHRWorkload} Prior computational studies have classified message types, identified patient concerns, and supported tasks such as message triage.\cite{Cronin2017MessageClassification,MerminBunnell2023CovidMessages,Ren2024PatientConcerns} EPPC focuses on a complementary problem: characterizing the communication behaviors expressed within a message and linking each classification to the Span that supports it.\cite{Fodeh2026EPPCMinerBen} This structure allows multiple communication behaviors to be represented within the same message and makes individual classifications directly reviewable. EPPC-OASIS previously automated this task by combining ontology-aware training with refinement of candidate triplets.\cite{Fodeh2026EPPCOASIS}

Earlier portal-message classification work identified communication types such as informational, medical, logistical, and social categories at the message level.\cite{Cronin2017MessageClassification} EPPC-CARE instead extracts individual communication behaviors as Span--Code--Sub-code triplets, allowing multiple behaviors within a message to be represented separately and linked to their supporting text.

The EPPC task also draws on broader work in clinical information extraction. Earlier approaches have included named-entity recognition, sequence labeling, and relation extraction, while pretrained biomedical language models and more recent instruction-following LLMs have expanded the types of structured information that can be extracted from clinical text.\cite{Nadkarni2011ClinicalNLP,Wang2018ClinicalIE,Kreimeyer2017ClinicalNLP,Lee2020BioBERT,Alsentzer2019ClinicalBERT,Agrawal2022ClinicalIE,Builtjes2025OpenIE} Several lines of work are particularly relevant to EPPC: Span extraction links predictions to source language, hierarchical classification represents relationships between Codes and Sub-codes, and structured generation supports a variable number of outputs.\cite{Geng2023Grammar,Lei2016Rationalizing,DeYoung2020ERASER,Silla2011HierarchicalSurvey,Mao2019HiLAP} EPPC requires all three capabilities because each triplet must include a Span together with its Code and Sub-code.

Retrieval and iterative processing provide additional strategies for complex language-model tasks. Retrieval can supply task-relevant examples at inference time, and prior work has shown that the choice of examples can affect model performance.\cite{Lewis2020RAG,Karpukhin2020DPR,Rubin2022RetrievePrompts} In EPPC-CARE, retrieval provides semantically similar annotated messages as examples of the communication-coding task rather than introducing external medical knowledge.\cite{Xiao2024BGE,Amugongo2025HealthcareRAG} Related work on iterative generation and refinement has also shown the value of separating complex outputs into successive steps that can use information from earlier decisions.\cite{Yao2023ReAct,Madaan2023SelfRefine,Shinn2023Reflexion,Jeong2024SelfBioRAG,Balachandran2025MedicalOrders} EPPC-CARE applies these ideas to patient--provider communication through Span identification, Code assignment, and Sub-code selection. Detailed implementation of this workflow is described in the Methods.

\section{Materials and Methods}\label{sec:methods}

\subsection{Study design}\label{sec:design}

We developed EPPC-CARE as a three-stage workflow for automated annotation of communication behaviors in secure patient--provider messages. For each message, EPPC-CARE identifies one or more Spans, assigns a Code to each Span, and then selects a corresponding Sub-code. Rather than requiring all three components of an annotation to be generated simultaneously, the workflow constructs each triplet sequentially through Span identification, Code assignment, and Sub-code selection.

We evaluated EPPC-CARE using the held-out EPPCMinerBen test set and compared it with conventional supervised fine-tuning (SFT) and EPPC-OASIS within each evaluated language model. Performance was assessed for Span identification, Code and Sub-code assignment, and triplets. To place model performance in the context of expert coding, we also compared EPPC-CARE and a human expert with consensus annotations on a random subset of 100 held-out messages.

\subsection{Dataset and annotation target}\label{sec:data}

We used the de-identified secure-message corpus and EPPC annotations introduced with the EPPCMinerBen benchmark.\cite{Fodeh2026EPPCMinerBen} The expanded, updated corpus contains 867 annotated messages from the patient portal of a single health system and 5,516 communication-behavior annotations. We retained the benchmark's stratified 70:30 partition of 607 training and 260 test messages. The annotations include 52 valid Code--Sub-code combinations across nine Codes and 34 Sub-codes. Each message also included its direction, indicating whether the message was directed toward the patient or toward the healthcare provider.

A message could contain multiple communication behaviors, each represented by a triplet consisting of a Span, Code, and Sub-code (Supplementary Table S1). We used the EPPCMinerBen consensus annotations (i.e., reference standard) without modification. Details of annotation development and adjudication are provided in Supplementary Methods.

\subsection{EPPC-CARE workflow}\label{sec:architecture}

\begin{figure}[t]
\centering
\includegraphics[width=\textwidth]{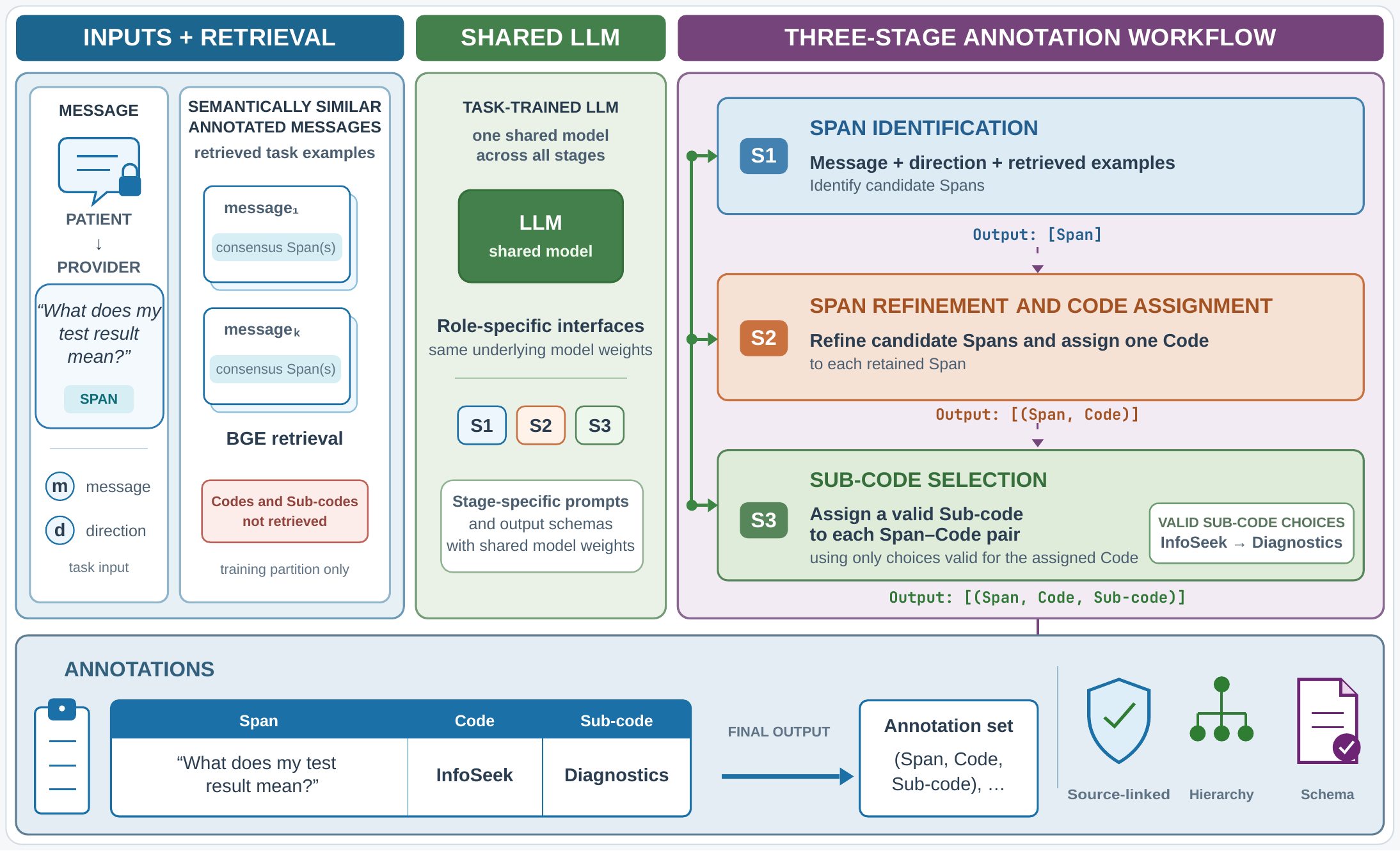}
\caption{EPPC-CARE staged workflow for communication-behavior annotation. For each secure patient--provider message, EPPC-CARE retrieves semantically similar annotated messages from the training set and constructs each annotation in three stages using a shared task-trained language model. BGE-large-en-v1.5 is the embedding model used to retrieve semantically similar training messages.\cite{Xiao2024BGE} Stage~1 identifies candidate Spans, Stage~2 refines the Spans and assigns Codes, and Stage~3 selects a Sub-code from the choices valid under the assigned Code. The resulting Span, Code, and Sub-code form a triplet that retains the source text supporting each classification.}
\label{fig:architecture}
\end{figure}

EPPC-CARE constructs each triplet through a fixed three-stage workflow (Figure~\ref{fig:architecture}). Before the first stage, the system retrieves semantically similar annotated messages from the training partition to provide examples of the EPPC coding task. A shared task-trained language model is then used sequentially for Span identification, Code assignment, and Sub-code selection. The output of each stage is passed to the next so that the annotation is constructed incrementally rather than generated in a single step.

Semantically similar examples were retrieved using a BGE-large-en-v1.5 embedding index constructed exclusively from the training partition.\cite{Xiao2024BGE} Detailed retrieval settings are provided in Supplementary Methods.

\subsubsection{Stage 1: Span identification}

Stage 1 received the message, its direction, and the retrieved annotated examples. The model identified candidate Spans corresponding to EPPC communication behaviors expressed in the message. These candidate Spans were passed to Stage 2.

\subsubsection{Stage 2: Span refinement and Code assignment}

Stage 2 received the original message, message direction, and candidate Spans from Stage 1. The model could add missed Spans, remove unsupported candidate Spans, or adjust Span boundaries before assigning a Code to each retained Span. The resulting Span--Code pairs were passed individually to Stage 3.

\subsubsection{Stage 3: Sub-code selection and annotation assembly}

For each Span--Code pair, Stage 3 received the original message, message direction, Span, assigned Code, and the Sub-codes valid under that Code in the EPPC hierarchy. The model selected a Sub-code from these valid choices. The implementation checked the selected Sub-code against the assigned Code's valid choices and corrected invalid selections before assembling the final triplet. The resulting Span, Code, and Sub-code formed a triplet, and all triplets for the message were assembled as the final output.

Decoding settings, output formatting, parsing procedures, and model-call accounting are provided in Supplementary Methods. Computational-efficiency analyses are reported separately in Supplementary Section S1.

\subsection{Task-specific model training}\label{sec:training}

The language models used in EPPC-CARE were trained to perform all three stages of the workflow. Training examples were converted into stage-specific instructions for Span identification, Span refinement and Code assignment, and Sub-code selection. Only examples from the training partition were used for this training. During end-to-end evaluation on the test set, each stage received predictions generated by the preceding stage rather than the reference standard.

Because Stage 2 and Stage 3 must operate on potentially imperfect predictions from earlier stages, their training inputs were perturbed to expose the models to errors similar to those that could occur during inference. The perturbation procedure is described in Supplementary Methods.

Each base model was trained in three independent runs using low-rank adaptation (LoRA) or quantized LoRA (QLoRA).\cite{Hu2022LoRA,Dettmers2023QLoRA} For the primary EPPC-CARE evaluation, corresponding adapter parameters from the three runs were averaged to produce one checkpoint for each language model and then merged with the base model for inference. This procedure combined trained model parameters rather than predictions or evaluation scores and was motivated by parameter-averaging approaches such as stochastic weight averaging.\cite{Izmailov2018} Complete training and optimization settings are reported in Supplementary Methods and Supplementary Table S3.

\subsection{Models and comparison systems}\label{sec:comparators}

We evaluated six instruction-tuned, open-weight language models ranging from 1B to 70B parameters: Llama-3.2-1B, Llama-3.2-3B, Llama-3.1-8B, Mistral-Small-24B, Qwen2.5-32B, and Llama-3.3-70B.\cite{MetaLlama31ModelCard,MetaLlama32ModelCard,MetaLlama33ModelCard,Mistral2025,QwenModelCard} These models represent three model families and a broad range of parameter sizes, allowing us to assess whether the effect of the EPPC-CARE workflow was consistent across different underlying models.

We compared three annotation approaches within the same corpus partitions and evaluation framework. SFT generated the Span, Code, and Sub-code together in a single pass. EPPC-OASIS used ontology-aware training and refinement of candidate triplets.\cite{Fodeh2026EPPCOASIS} EPPC-CARE instead constructed each triplet sequentially through Span identification, Code assignment, and Sub-code selection. These three approaches were compared across the 1B--70B models (Supplementary Table S2).

Previously reported prompting and preference-optimization results for Llama-3.1-8B were also considered as secondary benchmark context.\cite{Fodeh2026EPPCOASIS} These comparisons and their implementation details are described in the Supplement.

\subsection{EPPC-CARE component analysis}\label{sec:ablation-method}

We evaluated three modified versions of EPPC-CARE using the parameter-averaged Llama-3.2-3B checkpoint to determine which components of the workflow contributed to performance. The first omitted retrieval of semantically similar annotated messages. The second removed Span refinement in Stage 2, requiring the model to assign Codes directly to the Spans generated by Stage 1. The third removed the Code-based restriction in Stage 3 and allowed the model to choose from the full inventory of EPPC Sub-codes.

All variants used the same test examples, model checkpoint, decoding configuration, parser, and evaluation procedure. SFT provided the separately trained single-pass comparison.

\subsection{Evaluation metrics}\label{sec:metrics}

We evaluated performance using four relaxed-matching micro-F1 measures. Span F1 measured Span identification without considering Code or Sub-code assignment. Code+Span F1 required both a matching Span and the correct Code, whereas Sub-code+Span F1 required a matching Span and the correct Sub-code. Triplet F1 measured triplet performance by requiring agreement on the Span, Code, and Sub-code.

Predicted and reference annotations were matched one-to-one within each message using the prespecified relaxed Span-matching criterion. Normalization and matching procedures are provided in Supplementary Methods.

We also calculated macro-F1 to assess performance across communication categories without allowing frequently represented categories to dominate the overall score. Code+Span, Sub-code+Span, and Triplet F1 were calculated separately for each represented Code, Sub-code, or Code--Sub-code combination and then averaged across the nine Codes, 34 Sub-codes, and 48 Code--Sub-code pairs represented in the test set. SFT scores represent means across three independent training runs; EPPC-OASIS and EPPC-CARE scores were obtained from parameter-averaged checkpoints.

\subsection{Comparison with human expert annotation}\label{sec:expert-method}

To compare EPPC-CARE performance with human expert annotation, we randomly selected 100 messages from the held-out test set. One human expert from the research team independently annotated these messages using the same EPPC codebook without access to the reference standard. This independent comparison follows standard practice for evaluating clinical NLP reference standards.\cite{Ogren2006MedicalCorpora,Roberts2009GoldStandards} Agreement was evaluated for Span, Code, Sub-code, joint Code--Sub-code classification, and triplets using micro-F1. For each task, we calculated the difference between EPPC-CARE--reference standard agreement and human expert--reference standard agreement, with positive values indicating numerically closer agreement between EPPC-CARE and the reference standard. Confidence intervals for these paired differences were estimated by resampling the 100 messages. Direct agreement between EPPC-CARE and the human expert was also evaluated as a complementary measure.

\subsection{Reproducibility and privacy}\label{sec:implementation}

All model training, retrieval, inference, and evaluation were performed within a secure institutional computing environment using locally hosted open-weight language models. The description of the data, models, evaluation, and reproducibility provisions was informed by recommendations for reporting clinical AI and NLP studies.\cite{HernandezBoussard2020MINIMAR,GiladBachrach2025NLPReporting} The retrieval index contained only training-partition examples, and secure-message text remained within institutional infrastructure throughout the study. Source code, prompts, evaluation scripts, and trained model artifacts will be released through a public GitHub repository. Additional implementation and computing details are provided in the Supplement.

\section{Results}\label{sec:results}

\subsection{Performance Across Models}\label{sec:main-results}

\begin{figure}[t]
\centering
\includegraphics[width=\textwidth]{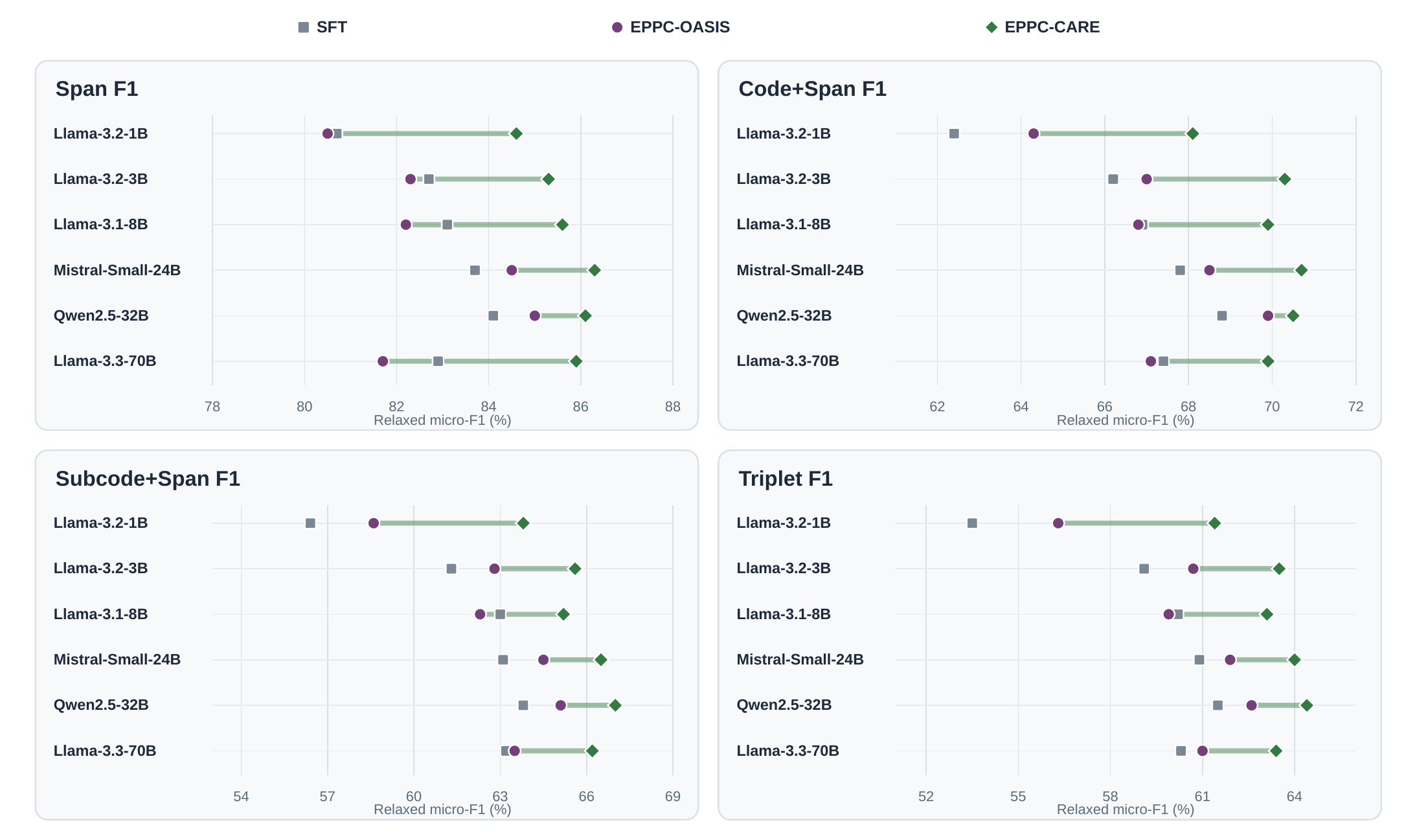}
\caption{Performance across language models on the 260-message held-out EPPC test set. Points show relaxed micro-F1 for SFT, EPPC-OASIS, and EPPC-CARE. Connecting segments show the difference between EPPC-OASIS and EPPC-CARE within each language model. Horizontal scales are specific to each metric.}
\label{fig:cross-backbone}
\end{figure}

EPPC-CARE outperformed all evaluated comparison methods across all six language models and all four micro-F1 measures (Figure~\ref{fig:cross-backbone} and Supplementary Table S4). Compared with the strongest comparison method, EPPC-OASIS, mean Triplet F1 increased by 2.90 percentage points across models, with individual EPPC-CARE scores ranging from 61.4\% to 64.4\%. Mean gains were 2.94 percentage points for Span F1, 2.64 for Code+Span F1, and 2.93 for Sub-code+Span F1. The improvement in Triplet F1 was maintained under alternative Span-matching and annotation-matching criteria (Supplementary Section S2). Precision and recall breakdowns comparing EPPC-CARE with SFT across all six models and seven extraction and label-only endpoints are reported in Supplementary Section~S3 (Supplementary Tables~S8--S10), showing that F1 gains were driven consistently by higher recall.

Performance gains were observed across model families and sizes rather than being limited to the largest language models. Qwen2.5-32B achieved the highest Triplet F1 at 64.4\%, whereas Mistral-Small-24B achieved the highest Span F1 (86.3\%) and Code+Span F1 (70.7\%). Llama-3.2-1B, the smallest evaluated model, achieved a Triplet F1 within 3.0 percentage points of the highest-performing model and improved by 5.1 points over its EPPC-OASIS counterpart. Thus, the benefit of the EPPC-CARE workflow was not dependent on a single model family or parameter scale.

Macro-F1 results indicated that the performance gains were not limited to the most frequently represented communication behaviors (Table~\ref{tab:macro-results}). EPPC-CARE achieved the highest macro-F1 across Code+Span, Sub-code+Span, and Triplet measures for five of the six evaluated models. Mistral-Small-24B was the exception: SFT achieved the highest Code+Span macro-F1 and EPPC-OASIS achieved the highest Sub-code+Span and Triplet macro-F1. These findings suggest that the overall EPPC-CARE advantage generally extended across the EPPC hierarchy, including less frequently represented behaviors.

\begin{table*}[t]
\caption{Performance on the 260-message held-out EPPC test set using relaxed macro-F1 (\%). Code+Span, Sub-code+Span, and Triplet scores are unweighted averages across the nine Codes, 34 Sub-codes, and 48 Code--Sub-code pairs represented in the test references, respectively. SFT values are means across three training runs; EPPC-OASIS and EPPC-CARE use parameter-averaged checkpoints.}
\label{tab:macro-results}
\centering\small
\begin{tabular}{@{}llrrr@{}}
\toprule
Model & Configuration & Code+Span & Sub-code+Span & Triplet \\
\midrule
Llama-3.2-1B
 & SFT & 45.01 & 39.22 & 40.18 \\
 & EPPC-OASIS & 50.72 & 41.05 & 42.76 \\
 & EPPC-CARE & \textbf{52.29} & \textbf{49.88} & \textbf{49.81} \\
\midrule
Llama-3.2-3B
 & SFT & 52.18 & 46.19 & 47.61 \\
 & EPPC-OASIS & 52.52 & 47.07 & 48.31 \\
 & EPPC-CARE & \textbf{53.74} & \textbf{50.23} & \textbf{50.23} \\
\midrule
Llama-3.1-8B
 & SFT & 52.32 & 49.30 & 49.61 \\
 & EPPC-OASIS & 49.66 & 48.06 & 48.18 \\
 & EPPC-CARE & \textbf{54.25} & \textbf{49.53} & \textbf{50.35} \\
\midrule
Mistral-Small-24B
 & SFT & \textbf{55.89} & 50.05 & 50.52 \\
 & EPPC-OASIS & 55.78 & \textbf{52.70} & \textbf{51.69} \\
 & EPPC-CARE & 54.64 & 49.90 & 50.13 \\
\midrule
Qwen2.5-32B
 & SFT & 56.27 & 49.30 & 49.56 \\
 & EPPC-OASIS & 58.41 & 50.03 & 50.69 \\
 & EPPC-CARE & \textbf{59.02} & \textbf{53.03} & \textbf{52.93} \\
\midrule
Llama-3.3-70B
 & SFT & 53.98 & 50.37 & 50.10 \\
 & EPPC-OASIS & 55.02 & 49.98 & 49.94 \\
 & EPPC-CARE & \textbf{56.55} & \textbf{52.36} & \textbf{51.77} \\
\bottomrule
\end{tabular}
\end{table*}

As secondary benchmark context, Table~\ref{tab:additional-baselines} compares the Llama-3.1-8B EPPC-CARE results with prompting, preference-optimization, and robustness-oriented methods previously evaluated on the EPPC benchmark. EPPC-CARE achieved higher Span and Triplet F1 than each additional baseline. Relative to the strongest of these methods, EPPC-CARE improved Span F1 by 2.24 percentage points over ORPO and Triplet F1 by 2.25 points over SimPO.

\begin{table}[t]
\caption{Additional benchmark context for Llama-3.1-8B using Span and Triplet micro-F1 (\%). Comparator results were reported in EPPC-OASIS.\cite{Fodeh2026EPPCOASIS}}
\label{tab:additional-baselines}
\centering\small
\begin{tabular}{@{}lrr@{}}
\toprule
Method & Span & Triplet \\
\midrule
Zero-shot prompting & 35.78 & 3.03 \\
Few-shot prompting & 75.56 & 16.29 \\
DPO & 83.25 & 60.81 \\
IPO & 83.29 & 60.58 \\
SimPO & 83.25 & 60.85 \\
ORPO & 83.36 & 60.72 \\
STaR-DRO & 82.95 & 59.69 \\
EPPC-CARE & \textbf{85.60} & \textbf{63.10} \\
\bottomrule
\end{tabular}
\end{table}

\subsection{Component analysis}\label{sec:ablation-results}

The largest performance contribution in the component analysis came from restricting Sub-code selection to choices valid under the assigned Code (Table~\ref{tab:ablations}). When Stage 3 was instead allowed to select from the full EPPC Sub-code inventory, Sub-code+Span F1 decreased by 2.22 percentage points and Triplet F1 decreased by 3.75 points. Removing retrieval of semantically similar annotated messages produced smaller reductions, while removing Span refinement in Stage 2 had little effect on aggregate performance. These results indicate that using the assigned Code to constrain subsequent Sub-code selection was the most influential workflow component evaluated.

\begin{table*}[t]
\caption{Component analysis using the parameter-averaged Llama-3.2-3B checkpoint on the 260-message held-out EPPC test set (relaxed micro-F1, \%). All configurations used the same checkpoint, decoding settings, and evaluator.}
\label{tab:ablations}
\centering\small
\begin{tabular}{@{}lrrrr@{}}
\toprule
Configuration & Span & Code+Span & Sub-code+Span & Triplet \\
\midrule
Full EPPC-CARE & \textbf{85.31} & 70.33 & \textbf{65.59} & \textbf{63.49} \\
Without retrieved annotated examples & 85.10 & \textbf{70.39} & 64.75 & 62.97 \\
Without Stage 2 Span refinement & 85.28 & 70.30 & 65.38 & 63.33 \\
Complete Sub-code inventory at Stage 3 & \textbf{85.31} & 70.33 & 63.37 & 59.74 \\
\bottomrule
\end{tabular}
\end{table*}

\subsection{Comparison with human expert annotation}\label{sec:expert-results}

\begin{figure}[t]
\centering
\includegraphics[width=\textwidth]{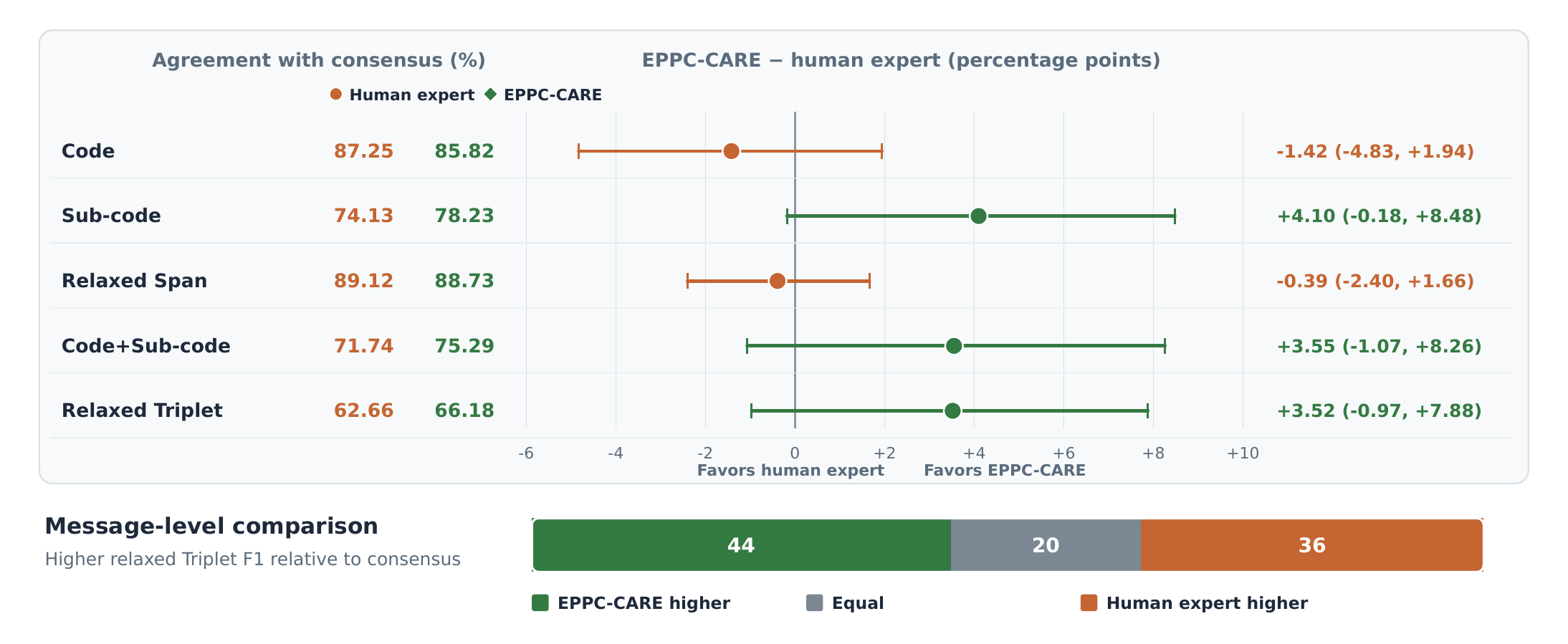}
\caption{Agreement of EPPC-CARE and a human expert with consensus annotations on 100 held-out messages. Differences are EPPC-CARE minus human expert agreement; error bars show paired bootstrap 95\% confidence intervals. The lower panel compares message-level relaxed Triplet F1 relative to the consensus reference.}
\label{fig:expert-comparison}
\end{figure}

On the 100-message human expert evaluation subset, EPPC-CARE and the human expert showed similar overall agreement with the reference standard, with the direction of the numerical difference varying across annotation measures (Figure~\ref{fig:expert-comparison} and Table~\ref{tab:expert}). EPPC-CARE had higher point estimates for Triplet F1 (66.18\% vs 62.66\%), Sub-code F1 (78.23\% vs 74.13\%), and joint Code--Sub-code F1 (75.29\% vs 71.74\%). The human expert had slightly higher point estimates for Code F1 (87.25\% vs 85.82\%) and Span F1 (89.12\% vs 88.73\%). The 95\% confidence intervals for all five differences included zero.

At the message level, EPPC-CARE achieved higher Triplet F1 than the human expert in 44 of the 100 messages, the human expert achieved higher Triplet F1 in 36, and performance was tied in 20. Direct Triplet agreement between EPPC-CARE and the human expert was 59.37\%. Together, these findings place EPPC-CARE's agreement with the consensus reference in the range observed for an independently annotating human expert on this evaluation subset, while showing different relative strengths across components of the EPPC annotation.

\begin{table*}[t]
\caption{Agreement on the 100-message human expert evaluation subset (micro-F1, \%). EPPC-CARE predictions were generated using Qwen2.5-32B. Differences compare EPPC-CARE--consensus with human expert--consensus agreement; positive values favor EPPC-CARE. Confidence intervals were calculated by paired bootstrap resampling.}
\label{tab:expert}
\centering\small
\setlength{\tabcolsep}{4.5pt}
\begin{tabular}{@{}lrrrr@{}}
\toprule
Metric
& \multicolumn{1}{r}{\shortstack[r]{Human expert\\vs consensus}}
& \multicolumn{1}{r}{\shortstack[r]{EPPC-CARE\\vs consensus}}
& \multicolumn{1}{r}{\shortstack[r]{Difference\\(95\% CI)}}
& \multicolumn{1}{r}{\shortstack[r]{EPPC-CARE\\vs human expert}} \\
\midrule
Code
& \textbf{87.25} & 85.82
& -1.42 (-4.83, 1.94) & 85.98 \\

Sub-code
& 74.13 & \textbf{78.23}
& +4.10 (-0.18, 8.48) & 72.69 \\

Relaxed Span
& \textbf{89.12} & 88.73
& -0.39 (-2.40, 1.66) & 89.31 \\

Code+Sub-code
& 71.74 & \textbf{75.29}
& +3.55 (-1.07, 8.26) & 69.83 \\

Relaxed Triplet
& 62.66 & \textbf{66.18}
& +3.52 (-0.97, 7.88) & 59.37 \\
\bottomrule
\end{tabular}
\end{table*}

\FloatBarrier
\section{Discussion}\label{sec:discussion}

\subsection{Principal findings}\label{sec:principal}

We developed EPPC-CARE to support scalable annotation of communication behaviors in secure patient--provider messages while retaining the Span associated with each classification. Across six language models spanning multiple families and sizes, EPPC-CARE consistently improved triplet performance compared with all evaluated comparison methods, with corresponding improvements in Span, Code+Span, and Sub-code+Span F1. These gains were generally maintained across less frequently represented communication behaviors, suggesting that the benefit was not confined to the most common EPPC categories.

EPPC-CARE outperformed both single-pass SFT and ontology-aware EPPC-OASIS. For Llama-3.1-8B, it also exceeded the previously reported prompting, preference-optimization, and robustness-oriented baselines, improving Span F1 by 2.24 percentage points over ORPO and Triplet F1 by 2.25 points over SimPO.\cite{Fodeh2026EPPCOASIS}

The human expert comparison provides additional context for interpreting these results. On the 100-message evaluation subset, EPPC-CARE and the human expert showed comparable agreement with the reference standard. Point estimates favored EPPC-CARE for Sub-code assignment, joint Code--Sub-code classification, and Triplet F1, whereas the human expert had slightly higher agreement for Span and Code annotation. The 95\% confidence intervals for all five differences included zero. These findings should not be interpreted as demonstrating replacement of expert annotation. Rather, they suggest that EPPC-CARE may be useful for generating initial annotations that can be reviewed and corrected by human experts, potentially extending the scale at which patient--provider communication can be studied. Such a role is consistent with human-in-the-loop approaches that use computational assistance to increase scale while retaining expert oversight.\cite{Wang2021HumanLoopNLP,Yuan2024HealthcareHITL}

This capability is relevant because secure-message corpora contain information about questions, requests, information exchange, care coordination, patient participation, and other communication behaviors that are difficult to capture through structured clinical data alone. Making these behaviors available for analysis at larger scale could support studies of how communication varies across patients, clinical settings, and time, and how communication patterns relate to healthcare processes and outcomes.

\subsection{Why the staged workflow may improve annotation}\label{sec:mechanism}

EPPC-CARE divides triplet annotation into three sequential tasks: Span identification, Span refinement and Code assignment, and Sub-code selection. This structure reduces the number of decisions that must be made simultaneously and allows information generated at one stage to guide the next. Semantically similar annotated messages provide examples of the coding task, while the EPPC hierarchy limits Sub-code selection to choices that are valid for the assigned Code.

The EPPC-CARE component analysis suggests that this Code-based restriction was the most influential of the evaluated workflow components. Allowing Stage 3 to select from the complete Sub-code inventory produced the largest reduction in triplet performance, whereas removing retrieval or Span refinement had smaller effects. This finding suggests that explicitly incorporating the relationship between Codes and Sub-codes can help the model distinguish among closely related communication behaviors.

The staged structure may also improve reviewability. Because Span, Code, and Sub-code decisions are produced sequentially rather than as an indivisible prediction, errors can be examined at the point at which they arise. For research annotation workflows, this creates opportunities for human experts to review individual components of an annotation rather than reassessing every message entirely from the beginning. This emphasis on traceability and expert oversight is consistent with recommended practices for clinically oriented machine-learning systems.\cite{HernandezBoussard2020MINIMAR,Yuan2024HealthcareHITL}

\subsection{Implications for clinical informatics}\label{sec:implications}

The primary application of EPPC-CARE is as annotation infrastructure for patient--provider communication research. Large secure-message collections are increasingly available for research, but detailed communication coding remains difficult to scale when every message requires manual annotation. EPPC-CARE could provide an initial set of triplets for human expert review, allowing reviewers to verify or correct the Span, Code, and Sub-code rather than create each annotation de novo. This approach could make larger longitudinal and multi-institutional studies of patient--provider communication more feasible while preserving a direct connection between each classification and the source language.

Such a workflow could also support more targeted use of human expertise. For example, review could be prioritized for uncommon Codes or Sub-codes, annotations for which different stages provide inconsistent information, or predictions associated with low confidence. The present study did not evaluate such a workflow prospectively, so reductions in annotation time or reviewer burden cannot yet be assumed. However, the structure of EPPC-CARE provides a basis for testing whether computational assistance can increase annotation throughput without compromising final dataset quality.

The broader methodological principle may also apply to other clinical information-extraction tasks in which text must first be located and then assigned increasingly specific labels or attributes. In these settings, separating Span identification from subsequent classification may provide more interpretable intermediate outputs and allow domain constraints to be incorporated at the point where they are most relevant. Whether the benefits observed for EPPC generalize to other clinical annotation tasks will require direct evaluation.

\subsection{Limitations}\label{sec:limitations}

This study has several limitations. First, the evaluation used messages from a single health system and an inherited train--test partition. Generalizability across institutions, patient populations, clinical specialties, and messaging practices therefore remains to be established.

Second, the human expert comparison involved one human expert and 100 held-out messages. This provides a useful reference point for interpreting model performance but does not characterize the full range of agreement that might be observed across multiple experts.

Third, EPPC-CARE uses multiple model calls and was evaluated primarily on annotation accuracy rather than in a prospective assisted-annotation workflow. Future studies should therefore assess the trade-off between performance, computational cost, and reviewer effort when the system is used at scale.

\subsection{Future directions}\label{sec:future}

Future studies should evaluate EPPC-CARE across independently constructed patient--provider messaging datasets from multiple institutions and prospectively compare fully manual coding with EPPC-CARE-assisted coding. Such studies can assess annotation time, correction burden, reviewer agreement, throughput, and final dataset quality in addition to annotation accuracy.

Additional work could examine confidence-based or disagreement-based review strategies that direct human experts toward uncertain Spans, Codes, or Sub-codes. If validated, these approaches could further concentrate expert effort on difficult annotations while allowing high-confidence cases to be processed more efficiently. Application to larger longitudinal corpora could then assess how scalable EPPC annotation changes the types and sizes of clinical communication studies that are feasible.

\section{Conclusion}\label{sec:conclusion}

EPPC-CARE provides a structured approach for automatically annotating communication behaviors in secure patient--provider messages while retaining the Span associated with each Code and Sub-code. The workflow improved triplet performance over all evaluated comparison methods across six language models, and its agreement with consensus annotations on the human expert evaluation subset was broadly comparable to that of one independently annotating human expert. These findings support further evaluation of EPPC-CARE as a tool for human-reviewed, large-scale communication annotation. Multi-institutional validation and prospective workflow studies are needed to determine its generalizability, efficiency, and effect on annotation quality.

\section*{Supplementary material}
The Appendix (Supplementary Material) accompanying this preprint provides supplementary methodological details, benchmark characteristics (Supplementary Table~S1), comparison-system configurations (Supplementary Table~S2), task-specific training settings (Supplementary Table~S3), complete cross-model micro-F1 results (Supplementary Table~S4), the matched inference-efficiency evaluation (Supplementary Section~S1, Table~S5), sensitivity analyses for Span matching, annotation assignment, and Span length (Supplementary Section~S2, Tables~S6--S7), and the cross-model precision and recall breakdown comparing EPPC-CARE against SFT (Supplementary Section~S3, Tables~S8--S10).

\section*{Ethics statement}
This study used an existing de-identified benchmark derived from secure patient--provider messages collected at Yale New Haven Hospital. The underlying secure-message data source was collected under approval of the Yale New Haven Hospital Institutional Review Board, as reported in prior work from this research program.\cite{Aantaki2023PCCSecureMessages} The present study was a secondary analysis of de-identified data; no direct patient identifiers were available to the investigators conducting this analysis. The data remained within the institutionally governed computing environment used for the EPPCMinerBen research program.

\section*{Data and code availability}
The clinical message corpus cannot be publicly released because of institutional privacy and data-governance restrictions. Source code, prompts, training configurations, and evaluation scripts will be made publicly available upon publication, without patient message text.

\section*{Funding}
This work was supported by the National Cancer Institute, United States (1R01CA285737-01A1).

\section*{Competing interests}
The authors declare that they have no known competing financial interests or personal relationships that could have appeared to influence the work reported in this paper.

\section*{Acknowledgments}
We thank Srivani Talakokkul and Afshan Khan for dataset annotation, data curation, codebook refinement, and quality assurance.

OpenAI Codex was used to assist with language editing and LaTeX formatting. The proposed changes were reviewed and approved by an author.

\section*{Author contributions}
\textbf{Samah Fodeh:} Conceptualization, Methodology, Resources, Supervision, Project administration, Writing -- original draft, Writing -- review and editing.

\textbf{Sreeraj Ramachandran:} Methodology, Software, Formal analysis, Investigation, Writing -- original draft, Writing -- review and editing.

\textbf{Ganesh Puthiaraju:} Validation, Writing -- review and editing.

\textbf{Sarah Schellhorn:} Validation, Writing -- review and editing.

\bibliographystyle{jamia}
\bibliography{references}

\clearpage
\appendix
\setcounter{section}{0}
\setcounter{table}{0}
\setcounter{figure}{0}
\renewcommand{\thesection}{S\arabic{section}}
\renewcommand{\thetable}{S\arabic{table}}
\renewcommand{\thefigure}{S\arabic{figure}}
\renewcommand{\theHsection}{S\arabic{section}}
\renewcommand{\theHtable}{S\arabic{table}}
\renewcommand{\theHfigure}{S\arabic{figure}}

\begin{center}
{\Large\bfseries Appendix / Supplementary Material}
\end{center}
\vspace{0.5em}

\section*{Supplementary Methods}

\subsection*{Annotation provenance}

The consensus annotations used in this study were inherited from EPPCMinerBen and were created using the structured EPPC codebook, which was informed by the Roter Interaction Analysis System.\cite{Roter2002RIAS,Fodeh2026EPPCMinerBen} Research-team annotators first completed a calibration phase to establish a shared interpretation of the EPPC Codes and Sub-codes. Two annotators then independently annotated each secure-message thread, and disagreements were resolved through adjudication and consensus meetings led by senior investigators. EPPCMinerBen reported inter-annotator agreement of Cohen's $\kappa=0.74$. The present study used the resulting consensus annotations without modification.

\subsection*{Retrieval of semantically similar annotated messages}

We constructed a dense retrieval index from the 607 training messages using BAAI BGE-large-en-v1.5.\cite{Xiao2024BGE} Message representations were obtained by tokenizing each message with padding and truncation to the encoder's 512-token limit, extracting the first token of the encoder's final hidden state, and applying $L_2$ normalization. No retrieval-specific query instruction was prepended.

Training messages were embedded in batches of 16, whereas each query message was embedded separately. Cosine similarity between normalized representations was used to identify the $k$ nearest annotated training messages. We evaluated retrieval of three to five examples on the validation set and used three examples in the final configuration. Each retrieved example contained at most the first 400 characters of the message together with its consensus Spans; Codes and Sub-codes were excluded from the retrieved examples. During construction of the task-specific training data, the source example was excluded from its own retrieval results by index. During test evaluation, retrieval was performed exclusively from the training partition.

\subsection*{Training-time perturbations for downstream stages}

Because Stage 2 and Stage 3 receive predictions from preceding stages during end-to-end inference, their training inputs were intentionally perturbed to expose the models to plausible upstream errors.

For Stage 2, inputs were generated from consensus Spans and perturbed before being paired with the complete consensus Span--Code targets. Each Span was independently omitted with probability 0.20. Surviving Spans were selected for boundary perturbation with probability 0.30. Selected Spans containing at least five characters were modified with probability 0.40 by deleting one to three characters from either boundary or extending the first exact occurrence in the source message by up to two characters at one boundary. Short Spans and unsuccessful extensions remained unchanged. In addition, a one- to three-word phrase sampled from the source message was appended with probability 0.20 when the message contained more than three whitespace-delimited words.

For Stage 3, the supplied Code was replaced with \texttt{InfoGive} with probability 0.15, except when the consensus Code was already \texttt{InfoGive}, in which case it was replaced with \texttt{InfoSeek}. The consensus Span--Code--Sub-code annotation remained the training target.

\subsection*{Training and optimization settings}

All models were trained with a maximum sequence length of 4,096, an effective batch size of 16, cosine learning-rate scheduling, a warm-up ratio of 0.10, and weight decay of 0.01. Optimization used 8-bit AdamW with bfloat16 computation.

Llama-3.2-1B, Llama-3.2-3B, and Llama-3.1-8B were trained for four epochs with a learning rate of $4\times10^{-4}$, LoRA rank $r=32$, and scaling coefficient $\alpha=64$. Mistral-Small-24B and Qwen2.5-32B were loaded in 4-bit precision using QLoRA.\cite{Dettmers2023QLoRA} Mistral-Small-24B was trained for three epochs with a learning rate of $2\times10^{-4}$, $r=32$, and $\alpha=64$; Qwen2.5-32B was trained for three epochs with a learning rate of $3\times10^{-4}$, $r=16$, and $\alpha=16$. Llama-3.3-70B was trained for two epochs with a learning rate of $1\times10^{-4}$, $r=16$, and $\alpha=32$ (Supplementary Table~S3).

\subsection*{Relaxed Span matching}

For relaxed matching, Spans were Unicode-case-folded, tokenized on whitespace, and represented as token sets. Two Spans were considered a match when either token set contained the other or their Jaccard similarity was at least 0.60. Predicted and reference annotations were matched one-to-one within each message. Code and Sub-code comparisons required exact, case-sensitive agreement. Counts were pooled across the test set to calculate micro-precision, micro-recall, and micro-F1.

\subsection*{Structured outputs and parsing}

Stage 1 generated the structured output \texttt{\{"spans": [span$_1$, $\ldots$, span$_n$]\}}. To recover structured predictions, the shared parser examined the complete model response, JSON-formatted Markdown fences, and candidate substrings delimited by braces or brackets, returning the first decodable JSON value. The recovered \texttt{spans} field was then supplied to Stage 2.

Stage 2 generated \texttt{\{"results": [\{"Span": ..., "Code": ...\}, ...]\}}. The parser accepted either the \texttt{results} field from a recovered JSON object or a directly generated list. The resulting Span--Code pairs were supplied individually to Stage 3 for Sub-code selection.

\subsection*{Decoding and model-call accounting}

All three stages used greedy decoding with temperature 0, a maximum of 1,024 generated tokens, and a model-specific end-of-turn stop token. The model context length was 4,096 tokens.

Each message required one Stage 1 call, one Stage 2 call, and one Stage 3 call for every Span--Code pair retained after Stage 2. The total number of model calls was therefore $2+n$, where $n$ was the number of Span--Code pairs passed to Stage 3. Inference cost consequently varied with the number of candidate annotations retained after Span refinement and Code assignment.

\subsection*{Task-specific training-instance construction}

Each corpus message contributed one Stage 1 Span-identification instance, one Stage 2 Span-refinement and Code-assignment instance, and one Stage 3 Sub-code-selection instance for each consensus annotation. The transformed training set contained 5,046 task-specific instances: 607 Stage 1 instances, 607 Stage 2 instances, and 3,832 Stage 3 instances. The corresponding transformed test set contained 2,204 instances: 260 Stage 1 instances, 260 Stage 2 instances, and 1,684 Stage 3 instances.

\FloatBarrier
\begin{table}[t]
\caption{Characteristics of the EPPC benchmark used for task-specific training and evaluation.}
\label{tab:dataset}
\centering
\begin{tabular}{@{}lr@{}}
\toprule
Characteristic & Value \\
\midrule
Annotated messages & 867 \\
Training / test messages & 607 / 260 \\
Communication-behavior annotations & 5,516 \\
Mean annotations per message & 6.36 \\
Codes & 9 \\
Sub-codes & 34 \\
Observed Code--Sub-code pairs & 52 \\
\bottomrule
\end{tabular}
\end{table}

\begin{table*}[t]
\caption{Annotation approaches compared within each language model.}
\label{tab:systems}
\centering\small
\begin{tabularx}{\textwidth}{@{}p{0.14\textwidth}p{0.31\textwidth}X@{}}
\toprule
Configuration & Task-specific training & Annotation process \\
\midrule
SFT
& Training to generate EPPC triplets
& Single-pass generation of the Span, Code, and Sub-code. \\

EPPC-OASIS
& Ontology-aware training with parameter averaging across three training runs
& Generation and structured refinement of candidate triplets using the averaged parameters. \\

EPPC-CARE
& Training for Span identification, Span refinement and Code assignment, and Sub-code selection; LoRA adapters averaged across three training runs
& Retrieval of semantically similar annotated messages followed by sequential construction of each triplet across three stages. \\
\bottomrule
\end{tabularx}
\end{table*}

\begin{table*}[t]
\caption{Model-specific EPPC-CARE training settings. Values apply to each of the three training seeds. All runs used a maximum sequence length of 4,096, cosine learning-rate scheduling, a warm-up ratio of 0.10, and weight decay of 0.01.}
\label{tab:adaptation}
\centering\small
\begin{tabular}{@{}lrrrrr@{}}
\toprule
Language model & Epochs & Learning rate & LoRA $r$ & LoRA $\alpha$ & Effective batch size \\
\midrule
Llama-3.2-1B      & 4 & $4\times10^{-4}$ & 32 & 64 & 16 \\
Llama-3.2-3B      & 4 & $4\times10^{-4}$ & 32 & 64 & 16 \\
Llama-3.1-8B      & 4 & $4\times10^{-4}$ & 32 & 64 & 16 \\
Mistral-Small-24B & 3 & $2\times10^{-4}$ & 32 & 64 & 16 \\
Qwen2.5-32B       & 3 & $3\times10^{-4}$ & 16 & 16 & 16 \\
Llama-3.3-70B     & 2 & $1\times10^{-4}$ & 16 & 32 & 16 \\
\bottomrule
\end{tabular}
\end{table*}
\begin{table*}[t]
\caption{Performance on the 260-message held-out EPPC test set using relaxed micro-F1 (\%). SFT values are means across three training runs; EPPC-OASIS and EPPC-CARE use parameter-averaged checkpoints. Triplet F1 measures joint agreement on Span, Code, and Sub-code.}
\label{tab:main-results}
\centering\small
\begin{tabular}{@{}llrrrr@{}}
\toprule
Model & Configuration & Span & Code+Span & Sub-code+Span & Triplet \\
\midrule
Llama-3.2-1B & SFT & 80.7 & 62.4 & 56.4 & 53.5 \\
 & EPPC-OASIS & 80.5 & 64.3 & 58.6 & 56.3 \\
 & EPPC-CARE & \textbf{84.6} & \textbf{68.1} & \textbf{63.8} & \textbf{61.4} \\
\midrule
Llama-3.2-3B & SFT & 82.7 & 66.2 & 61.3 & 59.1 \\
 & EPPC-OASIS & 82.3 & 67.0 & 62.8 & 60.7 \\
 & EPPC-CARE & \textbf{85.3} & \textbf{70.3} & \textbf{65.6} & \textbf{63.5} \\
\midrule
Llama-3.1-8B & SFT & 83.1 & 66.9 & 63.0 & 60.2 \\
 & EPPC-OASIS & 82.2 & 66.8 & 62.3 & 59.9 \\
 & EPPC-CARE & \textbf{85.6} & \textbf{69.9} & \textbf{65.2} & \textbf{63.1} \\
\midrule
Mistral-Small-24B & SFT & 83.7 & 67.8 & 63.1 & 60.9 \\
 & EPPC-OASIS & 84.5 & 68.5 & 64.5 & 61.9 \\
 & EPPC-CARE & \textbf{86.3} & \textbf{70.7} & \textbf{66.5} & \textbf{64.0} \\
\midrule
Qwen2.5-32B & SFT & 84.1 & 68.8 & 63.8 & 61.5 \\
 & EPPC-OASIS & 85.0 & 69.9 & 65.1 & 62.6 \\
 & EPPC-CARE & \textbf{86.1} & \textbf{70.5} & \textbf{67.0} & \textbf{64.4} \\
\midrule
Llama-3.3-70B & SFT & 82.9 & 67.4 & 63.2 & 60.3 \\
  & EPPC-OASIS & 81.7 & 67.1 & 63.5 & 61.0 \\
 & EPPC-CARE & \textbf{85.9} & \textbf{69.9} & \textbf{66.2} & \textbf{63.4} \\
\midrule
\multicolumn{2}{l}{Mean EPPC-CARE gain over EPPC-OASIS} & +2.94 & +2.64 & +2.93 & +2.90 \\
\bottomrule
\end{tabular}
\end{table*}

\FloatBarrier
\section{Inference efficiency}\label{app:efficiency}

We characterized the computational cost of EPPC-CARE in a matched comparison with EPPC-OASIS using Llama-3.2-3B. Both approaches were evaluated with vLLM on a single NVIDIA B200 using greedy decoding, a maximum context length of 4,096 tokens, and prefix caching disabled.\cite{Kwon2023vLLM} After an eight-message warm-up, each approach processed the complete 260-message held-out test set three times. Checkpoint loading and, for EPPC-CARE, construction of the retrieval index were excluded from the timed interval. EPPC-CARE timing included retrieval of semantically similar annotated messages, prompt construction, all model calls, response parsing, and assembly of the triplets.

\begin{table*}[t]
\caption{Matched inference-efficiency comparison using Llama-3.2-3B on an NVIDIA B200. Values are means across three complete passes over the 260-message held-out test set. Amortized time reflects batched test-set processing and should not be interpreted as latency for an individual message. Peak device memory includes the key--value cache preallocated by vLLM at a GPU-memory utilization setting of 0.45.}
\label{tab:efficiency}
\centering\small
\begin{tabular}{@{}lrrr@{}}
\toprule
Measure & EPPC-OASIS & EPPC-CARE & CARE/OASIS \\
\midrule
Test-set time, s & 4.81 & 11.33 & 2.36 \\
Amortized time, ms/message & 18.50 & 43.58 & 2.36 \\
Throughput, messages/s & 54.08 & 22.95 & 0.42 \\
Model calls/message & 1.00 & 8.63 & 8.63 \\
Prompt tokens/message & 81.7 & 1,996.3 & 24.43 \\
Generated tokens/message & 210.7 & 449.1 & 2.13 \\
Peak device memory, GiB & 84.1 & 86.4 & 1.03 \\
\bottomrule
\end{tabular}
\end{table*}

EPPC-CARE required 2.36 times the wall-clock time of EPPC-OASIS and processed 22.95 compared with 54.08 messages per second. The increase in processing time was substantially smaller than the 8.63-fold increase in model calls because requests within each stage were processed in batches. Stage~3 accounted for an average of 6.63 calls per message because Sub-code selection was performed separately for each Span--Code pair retained after Stage~2. EPPC-CARE also processed substantially more prompt context because the original message and relevant output from the preceding stage were provided at successive stages. Generated-token volume increased 2.13-fold, whereas peak device-memory use increased by approximately 2.3~GiB. The absolute memory values largely reflect the preallocated key--value cache and are therefore specific to this runtime configuration. Across all four micro-F1 measures, performance during the efficiency evaluation differed by no more than 0.26 percentage points from the primary results.

\FloatBarrier
\section{Sensitivity to Span-matching criteria}\label{app:metric-sensitivity}

We examined whether the performance difference between EPPC-CARE and EPPC-OASIS depended on the Span-matching criterion used in the primary evaluation. Predictions from both configurations were evaluated using normalized exact equality and token-set Jaccard thresholds of 0.50, 0.60, and 0.70 without the containment shortcut. We also compared the order-dependent greedy assignment with maximum-cardinality bipartite matching while retaining the primary Span-compatibility criterion. Table~\ref{tab:metric-sensitivity} reports Triplet micro-F1 averaged across the included language models.

\begin{table*}[t]
\caption{Sensitivity of Triplet micro-F1 (\%) to Span matching and annotation assignment. Values are unweighted means across the included language models. Scores were recomputed at prediction level and may differ slightly from the aggregate values in the primary-results table (Supplementary Table S4).}
\label{tab:metric-sensitivity}
\centering\small
\begin{tabular}{@{}lrrr@{}}
\toprule
Matching and assignment rule & EPPC-OASIS & EPPC-CARE & Difference \\
\midrule
Primary relaxed matching; greedy assignment & 60.41 & 63.27 & +2.86 \\
Primary relaxed matching; maximum-cardinality assignment & 60.41 & 63.27 & +2.86 \\
Normalized exact Span equality & 35.81 & 37.64 & +1.83 \\
Token-set Jaccard $\geq 0.50$ & 56.08 & 59.48 & +3.40 \\
Token-set Jaccard $\geq 0.60$ & 50.76 & 53.61 & +2.84 \\
Token-set Jaccard $\geq 0.70$ & 45.28 & 47.85 & +2.57 \\
\bottomrule
\end{tabular}
\end{table*}

EPPC-CARE retained a positive Triplet F1 difference for every language model under every matching criterion. Replacing greedy assignment with maximum-cardinality matching changed any individual score by at most 0.06 percentage points, indicating that annotation ordering had negligible influence on the comparison. The advantage also remained under normalized exact Span equality, although the lower absolute scores under this criterion reflect the sensitivity of triplet performance to differences in Span boundaries.

We further stratified the primary Triplet evaluation by Span length in whitespace-delimited tokens (Table~\ref{tab:span-length-sensitivity}). References and predictions were assigned to strata according to their own normalized Span length and evaluated within each stratum. EPPC-CARE improved mean Triplet F1 for Spans of two or more tokens, with the largest difference observed for Spans containing two to three tokens. In contrast, its mean performance was 1.43 points lower for single-token Spans, identifying very short Span identification as a potential area for further improvement.

\begin{table*}[t]
\caption{Triplet micro-F1 (\%) stratified by Span length. Values are unweighted means across the included language models; support denotes the number of reference annotations in each stratum.}
\label{tab:span-length-sensitivity}
\centering\small
\begin{tabular}{@{}lrrrr@{}}
\toprule
Span length, tokens & Reference support & EPPC-OASIS & EPPC-CARE & Difference \\
\midrule
1 & 215 & 78.79 & 77.36 & -1.43 \\
2--3 & 538 & 42.10 & 45.14 & +3.03 \\
4--7 & 840 & 43.47 & 46.20 & +2.73 \\
$\geq 8$ & 91 & 23.68 & 25.71 & +2.02 \\
\bottomrule
\end{tabular}
\end{table*}

\FloatBarrier
\section{Precision and recall breakdown: EPPC-CARE versus SFT}\label{app:precision-recall}

To examine whether the F1 gains of EPPC-CARE over single-pass supervised fine-tuning (SFT) were driven by precision, recall, or both, we recomputed micro-precision, micro-recall, and micro-F1 uniformly from the saved prediction artifacts on the same 260-message held-out EPPC test set across all six language models. SFT metrics represent the arithmetic mean across three independent training seeds, whereas EPPC-CARE uses each model's parameter-averaged checkpoint. In addition to the four primary span-dependent extraction endpoints evaluated with relaxed Span matching (Triplet, Span, Code+Span, and Sub-code+Span), we evaluated three label-only endpoints (Code, Sub-code, and Code+Sub-code) in which Spans are ignored and each distinct label or pair is counted once per message. Because all baselines were rescored uniformly at the prediction level with the audited evaluator (including the 70B SFT re-evaluation artifacts), some SFT F1 values differ slightly from the aggregate values in Supplementary Table~S4.

Across all 42 model--endpoint comparisons (six models $\times$ seven endpoints), EPPC-CARE improved both micro-recall (42/42) and micro-F1 (42/42) relative to SFT, whereas precision increased in 15 comparisons and decreased in 27. The observed F1 improvements are therefore driven consistently by higher recall rather than uniform gains in precision. For the primary relaxed Triplet endpoint (Table~\ref{tab:pr-triplet}), mean recall increased by $+7.26$ percentage points (range $+4.73$ to $+11.54$), whereas mean precision changed by $+0.68$ percentage points (range $-1.66$ to $+3.92$), yielding a mean Triplet F1 gain of $+4.04$ percentage points (range $+2.69$ to $+7.84$). The largest simultaneous gains occurred on the smallest backbone, Llama-3.2-1B ($+3.92$ points in precision, $+11.54$ in recall, and $+7.84$ in F1), while larger models (8B--70B) achieved $+4.73$ to $+6.93$ percentage-point recall gains with smaller precision shifts ($-1.66$ to $+0.63$ points).

\begin{table*}[t]
\caption{Relaxed Triplet micro-precision, micro-recall, and micro-F1 (\%) for SFT and EPPC-CARE on the 260-message held-out EPPC test set across all six language models. SFT values are arithmetic means across three independent training runs; EPPC-CARE uses the parameter-averaged checkpoint. Differences ($\Delta$, EPPC-CARE minus SFT in percentage points) were computed before rounding. Bold indicates the higher score within each SFT--EPPC-CARE pair. Scores were recomputed uniformly from saved prediction artifacts and may differ slightly from the aggregate SFT values in Supplementary Table~S4.}
\label{tab:pr-triplet}
\centering\footnotesize
\setlength{\tabcolsep}{4pt}
\begin{tabular}{@{}lrrrrrrrrr@{}}
\toprule
 & \multicolumn{3}{c}{Precision (\%)} & \multicolumn{3}{c}{Recall (\%)} & \multicolumn{3}{c}{F1 (\%)} \\
\cmidrule(lr){2-4}\cmidrule(lr){5-7}\cmidrule(lr){8-10}
Model & SFT & CARE & $\Delta$ & SFT & CARE & $\Delta$ & SFT & CARE & $\Delta$ \\
\midrule
Llama-3.2-1B & 56.18 & \textbf{60.10} & $+$3.92 & 51.17 & \textbf{62.71} & $+$11.54 & 53.54 & \textbf{61.38} & $+$7.84 \\
Llama-3.2-3B & 61.08 & \textbf{62.57} & $+$1.49 & 56.63 & \textbf{64.43} & $+$7.80 & 58.77 & \textbf{63.49} & $+$4.72 \\
Llama-3.1-8B & \textbf{63.30} & 61.64 & $-$1.66 & 57.68 & \textbf{64.61} & $+$6.93 & 60.36 & \textbf{63.09} & $+$2.73 \\
Mistral-Small-24B & \textbf{63.16} & 62.80 & $-$0.36 & 58.75 & \textbf{65.26} & $+$6.51 & 60.87 & \textbf{64.01} & $+$3.14 \\
Qwen2.5-32B & 62.94 & \textbf{63.56} & $+$0.63 & 60.53 & \textbf{65.26} & $+$4.73 & 61.71 & \textbf{64.40} & $+$2.69 \\
Llama-3.3-70B & 62.19 & \textbf{62.26} & $+$0.07 & 58.47 & \textbf{64.55} & $+$6.08 & 60.27 & \textbf{63.38} & $+$3.12 \\
\midrule
Mean across models & 61.47 & \textbf{62.16} & $+$0.68 & 57.21 & \textbf{64.47} & $+$7.26 & 59.25 & \textbf{63.29} & $+$4.04 \\
\bottomrule
\end{tabular}
\end{table*}

The same recall-driven pattern appeared across the remaining span-dependent endpoints (Table~\ref{tab:pr-span-endpoints}). For Span identification, EPPC-CARE increased recall for every model by $+4.75$ to $+9.28$ percentage points (mean $+7.26$) while modestly lowering precision by $-0.80$ to $-3.76$ percentage points (mean $-1.86$), resulting in net Span F1 gains of $+2.00$ to $+3.89$ points (mean $+2.80$). As label specificity increased from Span to Code+Span and Sub-code+Span, mean precision shifted from $-1.86$ to $-0.43$ and $+0.47$ percentage points while mean recall gains remained large ($+6.93$ and $+7.32$ points), reflecting the additional benefit of hierarchy-constrained Code and Sub-code assignment in Stages~2 and~3.

\begin{table*}[t]
\caption{Micro-precision, micro-recall, and micro-F1 (\%) for SFT and EPPC-CARE across the remaining span-dependent extraction endpoints (Span, Code+Span, and Sub-code+Span) on the 260-message held-out EPPC test set. SFT values are means across three training runs; EPPC-CARE uses the parameter-averaged checkpoint. Differences ($\Delta$, EPPC-CARE minus SFT in percentage points) were computed before rounding. Bold indicates the higher score within each SFT--EPPC-CARE pair.}
\label{tab:pr-span-endpoints}
\centering\scriptsize
\setlength{\tabcolsep}{3.5pt}
\begin{tabular}{@{}llrrrrrrrrr@{}}
\toprule
 & & \multicolumn{3}{c}{Precision (\%)} & \multicolumn{3}{c}{Recall (\%)} & \multicolumn{3}{c}{F1 (\%)} \\
\cmidrule(lr){3-5}\cmidrule(lr){6-8}\cmidrule(lr){9-11}
Model & Endpoint & SFT & CARE & $\Delta$ & SFT & CARE & $\Delta$ & SFT & CARE & $\Delta$ \\
\midrule
Llama-3.2-1B & Span & \textbf{84.66} & 82.81 & $-$1.85 & 77.12 & \textbf{86.40} & $+$9.28 & 80.68 & \textbf{84.57} & $+$3.89 \\
 & Code+Span & 65.45 & \textbf{66.70} & $+$1.26 & 59.62 & \textbf{69.60} & $+$9.98 & 62.37 & \textbf{68.12} & $+$5.75 \\
 & Sub-code+Span & 59.18 & \textbf{62.44} & $+$3.25 & 53.90 & \textbf{65.14} & $+$11.24 & 56.40 & \textbf{63.76} & $+$7.36 \\
\midrule
Llama-3.2-3B & Span & \textbf{85.57} & 84.08 & $-$1.49 & 79.33 & \textbf{86.58} & $+$7.24 & 82.33 & \textbf{85.31} & $+$2.98 \\
 & Code+Span & 68.41 & \textbf{69.32} & $+$0.91 & 63.42 & \textbf{71.38} & $+$7.96 & 65.82 & \textbf{70.33} & $+$4.52 \\
 & Sub-code+Span & 63.38 & \textbf{64.65} & $+$1.26 & 58.77 & \textbf{66.57} & $+$7.80 & 60.99 & \textbf{65.59} & $+$4.61 \\
\midrule
Llama-3.1-8B & Span & \textbf{87.38} & 83.63 & $-$3.76 & 79.63 & \textbf{87.65} & $+$8.02 & 83.32 & \textbf{85.59} & $+$2.27 \\
 & Code+Span & \textbf{70.23} & 68.27 & $-$1.96 & 63.99 & \textbf{71.56} & $+$7.56 & 66.96 & \textbf{69.88} & $+$2.91 \\
 & Sub-code+Span & \textbf{66.04} & 63.74 & $-$2.30 & 60.17 & \textbf{66.81} & $+$6.63 & 62.97 & \textbf{65.24} & $+$2.27 \\
\midrule
Mistral-Small-24B & Span & \textbf{86.80} & 84.63 & $-$2.17 & 80.74 & \textbf{87.95} & $+$7.21 & 83.66 & \textbf{86.26} & $+$2.60 \\
 & Code+Span & \textbf{70.35} & 69.37 & $-$0.98 & 65.44 & \textbf{72.09} & $+$6.65 & 67.80 & \textbf{70.70} & $+$2.90 \\
 & Sub-code+Span & \textbf{65.50} & 65.26 & $-$0.24 & 60.93 & \textbf{67.81} & $+$6.89 & 63.13 & \textbf{66.51} & $+$3.38 \\
\midrule
Qwen2.5-32B & Span & \textbf{85.76} & 84.96 & $-$0.80 & 82.48 & \textbf{87.23} & $+$4.75 & 84.09 & \textbf{86.08} & $+$2.00 \\
 & Code+Span & \textbf{70.49} & 69.58 & $-$0.92 & 67.79 & \textbf{71.44} & $+$3.64 & 69.12 & \textbf{70.50} & $+$1.38 \\
 & Sub-code+Span & 65.18 & \textbf{66.17} & $+$0.99 & 62.69 & \textbf{67.93} & $+$5.25 & 63.91 & \textbf{67.04} & $+$3.13 \\
\midrule
Llama-3.3-70B & Span & \textbf{85.53} & 84.42 & $-$1.11 & 80.44 & \textbf{87.53} & $+$7.09 & 82.90 & \textbf{85.95} & $+$3.05 \\
 & Code+Span & \textbf{69.54} & 68.67 & $-$0.87 & 65.40 & \textbf{71.20} & $+$5.80 & 67.40 & \textbf{69.91} & $+$2.51 \\
 & Sub-code+Span & \textbf{65.23} & 65.06 & $-$0.17 & 61.34 & \textbf{67.46} & $+$6.12 & 63.22 & \textbf{66.24} & $+$3.02 \\
\midrule
Mean across models & Span & \textbf{85.95} & 84.09 & $-$1.86 & 79.96 & \textbf{87.22} & $+$7.26 & 82.83 & \textbf{85.63} & $+$2.80 \\
 & Code+Span & \textbf{69.08} & 68.65 & $-$0.43 & 64.28 & \textbf{71.21} & $+$6.93 & 66.58 & \textbf{69.91} & $+$3.33 \\
 & Sub-code+Span & 64.09 & \textbf{64.55} & $+$0.47 & 59.63 & \textbf{66.95} & $+$7.32 & 61.77 & \textbf{65.73} & $+$3.96 \\
\bottomrule
\end{tabular}
\end{table*}

Label-only (span-agnostic) results confirm that recall gains are not solely an artifact of Span extraction (Table~\ref{tab:pr-label-endpoints}). Ignoring Spans, EPPC-CARE still improved recall across every model for Code (mean $+2.47$ points), Sub-code (mean $+4.81$ points), and Code+Sub-code (mean $+4.81$ points), with small mean precision changes ($-0.06$, $-0.71$, and $-0.53$ points, respectively), producing net F1 gains of $+1.30$, $+2.19$, and $+2.28$ percentage points. Consistent with the component analysis in the main text, gains were larger for fine-grained Sub-code and joint Code+Sub-code recovery than for top-level Code assignment.

\begin{table*}[t]
\caption{Label-only (span-agnostic) micro-precision, micro-recall, and micro-F1 (\%) for SFT and EPPC-CARE on the 260-message held-out EPPC test set. Spans are ignored and each distinct Code, Sub-code, or Code--Sub-code pair is counted once per message. SFT values are means across three training runs; EPPC-CARE uses the parameter-averaged checkpoint. Differences ($\Delta$, EPPC-CARE minus SFT in percentage points) were computed before rounding. Bold indicates the higher score within each SFT--EPPC-CARE pair.}
\label{tab:pr-label-endpoints}
\centering\scriptsize
\setlength{\tabcolsep}{3.5pt}
\begin{tabular}{@{}llrrrrrrrrr@{}}
\toprule
 & & \multicolumn{3}{c}{Precision (\%)} & \multicolumn{3}{c}{Recall (\%)} & \multicolumn{3}{c}{F1 (\%)} \\
\cmidrule(lr){3-5}\cmidrule(lr){6-8}\cmidrule(lr){9-11}
Model & Endpoint & SFT & CARE & $\Delta$ & SFT & CARE & $\Delta$ & SFT & CARE & $\Delta$ \\
\midrule
Llama-3.2-1B & Code & \textbf{88.62} & 85.95 & $-$2.66 & 77.84 & \textbf{83.53} & $+$5.69 & 82.86 & \textbf{84.73} & $+$1.87 \\
 & Sub-code & \textbf{78.90} & 78.49 & $-$0.41 & 70.06 & \textbf{78.42} & $+$8.36 & 74.21 & \textbf{78.46} & $+$4.25 \\
 & Code+Sub-code & \textbf{75.06} & 74.85 & $-$0.21 & 66.58 & \textbf{74.85} & $+$8.27 & 70.56 & \textbf{74.85} & $+$4.29 \\
\midrule
Llama-3.2-3B & Code & 88.70 & \textbf{90.19} & $+$1.49 & 81.36 & \textbf{84.57} & $+$3.21 & 84.86 & \textbf{87.29} & $+$2.43 \\
 & Sub-code & \textbf{80.54} & 80.31 & $-$0.23 & 74.21 & \textbf{78.33} & $+$4.12 & 77.24 & \textbf{79.31} & $+$2.07 \\
 & Code+Sub-code & 77.51 & \textbf{77.65} & $+$0.15 & 70.77 & \textbf{75.36} & $+$4.59 & 73.98 & \textbf{76.49} & $+$2.51 \\
\midrule
Llama-3.1-8B & Code & 88.34 & \textbf{89.33} & $+$0.99 & 81.31 & \textbf{83.23} & $+$1.93 & 84.67 & \textbf{86.18} & $+$1.50 \\
 & Sub-code & \textbf{82.00} & 79.61 & $-$2.39 & 74.21 & \textbf{78.42} & $+$4.21 & 77.91 & \textbf{79.01} & $+$1.10 \\
 & Code+Sub-code & \textbf{78.38} & 76.59 & $-$1.79 & 70.97 & \textbf{75.11} & $+$4.14 & 74.49 & \textbf{75.84} & $+$1.35 \\
\midrule
Mistral-Small-24B & Code & 88.27 & \textbf{89.50} & $+$1.23 & 84.08 & \textbf{84.72} & $+$0.64 & 86.12 & \textbf{87.04} & $+$0.92 \\
 & Sub-code & 80.50 & \textbf{81.35} & $+$0.85 & 75.56 & \textbf{80.35} & $+$4.80 & 77.94 & \textbf{80.85} & $+$2.91 \\
 & Code+Sub-code & 77.71 & \textbf{78.02} & $+$0.30 & 72.41 & \textbf{76.96} & $+$4.56 & 74.96 & \textbf{77.49} & $+$2.53 \\
\midrule
Qwen2.5-32B & Code & \textbf{88.94} & 87.78 & $-$1.15 & 85.06 & \textbf{86.35} & $+$1.29 & 86.95 & \textbf{87.06} & $+$0.11 \\
 & Sub-code & \textbf{80.83} & 80.40 & $-$0.43 & 76.70 & \textbf{80.61} & $+$3.92 & 78.71 & \textbf{80.51} & $+$1.80 \\
 & Code+Sub-code & \textbf{77.87} & 77.36 & $-$0.50 & 73.73 & \textbf{77.30} & $+$3.57 & 75.74 & \textbf{77.33} & $+$1.59 \\
\midrule
Llama-3.3-70B & Code & \textbf{87.69} & 87.46 & $-$0.23 & 82.79 & \textbf{84.87} & $+$2.08 & 85.17 & \textbf{86.14} & $+$0.97 \\
 & Sub-code & \textbf{81.32} & 79.68 & $-$1.64 & 75.99 & \textbf{79.47} & $+$3.48 & 78.57 & \textbf{79.58} & $+$1.01 \\
 & Code+Sub-code & \textbf{77.47} & 76.34 & $-$1.13 & 72.24 & \textbf{75.95} & $+$3.71 & 74.76 & \textbf{76.14} & $+$1.38 \\
\midrule
Mean across models & Code & \textbf{88.43} & 88.37 & $-$0.06 & 82.07 & \textbf{84.55} & $+$2.47 & 85.11 & \textbf{86.41} & $+$1.30 \\
 & Sub-code & \textbf{80.68} & 79.97 & $-$0.71 & 74.45 & \textbf{79.27} & $+$4.81 & 77.43 & \textbf{79.62} & $+$2.19 \\
 & Code+Sub-code & \textbf{77.33} & 76.80 & $-$0.53 & 71.12 & \textbf{75.92} & $+$4.81 & 74.08 & \textbf{76.36} & $+$2.28 \\
\bottomrule
\end{tabular}
\end{table*}

\end{document}